# Self-Organizing Map Neural Network Algorithm for the Determination of Fracture Location in Solid-State Process joined Dissimilar Alloys

Akshansh Mishra[1], Anish Dasgupta[2]

[1]School of Industrial and Information Engineering, Politecnico Di Milano, Milan, Italy

[2]Artificial Intelligence Analytics, Cognizant Technology Solutions, India

**Abstract:** The subject area known as computational neuroscience involves the investigation of brain function using mathematical techniques and theories. In order to comprehend how the brain processes information, it can also include various methods from signal processing, computer science, and physics. In the present work, for first time a neurobiological based unsupervised machine learning algorithm i.e., Self-Organizing Map Neural Network is implemented for determining the fracture location in dissimilar friction stir welded AA5754-C11000 alloys. Too Shoulder Diameter (mm), Tool Rotational Speed (RPM), and Tool Traverse Speed (mm/min) are input parameters while the Fracture location i.e., whether the specimen's fracture at Thermo-Mechanically Affected Zone (TMAZ) of copper or it fractures at TMAZ of Aluminium. The results showed that the implemented algorithm is able to predict the fracture location with 96.92% accuracy.

1. Introduction

Information processing carried out by networks of neurons is referred to as neural computing. The philosophical movement known as computational mind theory or computationalism, which promotes the idea that neural computation accounts cognition, has ties to neural computation [1-4]. Nowadays, these types of algorithms are used in manufacturing and materials sectors for the determination of mechanical and microstructure properties of fabricated alloys or specimens [5-6]. An artificial neural network (ANN) was used by Shiau et al. [7] to model Taiwan's industrial energy demand in relation to subsector industrial output and climate change. It was the first investigation to measure the relationship between industrial energy use, manufacturing output, and climate change using the ANN technique. A multilayer perceptron (MLP) with a feedforward backpropagation neural network was used as the ANN model in this investigation. In order to improve the implementation of natural fibers in green bio-composites, Jarrah et al. [8] used doubly interconnected artificial neural networks to make unique classifications and prediction of the inherent mechanical properties of natural fibers. Whether CNNs are effective for identifying internal weld faults in addition to surface defects was determined by Hartl et al. [9]. Ultrasonic testing was used to create 120 welds for this purpose, and that data was used to determine if it was "good" or "defective." Different artificial neural network models were examined for their ability to anticipate where the welds would fall within the designated classes. The method used to label the data was found to be important for the level of precision that could be attained.

In the present work, Self-Organizing Map Neural Network model is used for the first time for fracture location classification purposes in the Friction Stir Welding process.

## 2. Methodology

Figure 1 shows the process flow chart of the methodology implemented in the present research work. Firstly, the data is obtained from the experimental test [10] and is further prepared in the form of a CSV file where input parameters and output parameters are entered in column-wise format. In the present study, the tensile specimen, Tool Traverse Speed (mm/min), Tool Shoulder Diameter (mm), and Tool Rotational Speed (RPM) are the input parameters while the Fracture location is an output parameter. If the tensile specimen fractures at the TMAZ Cu zone, then its value is assigned as '0' and if the tensile specimen fractures at the TMAZ Al zone then its value is assigned as '1'.

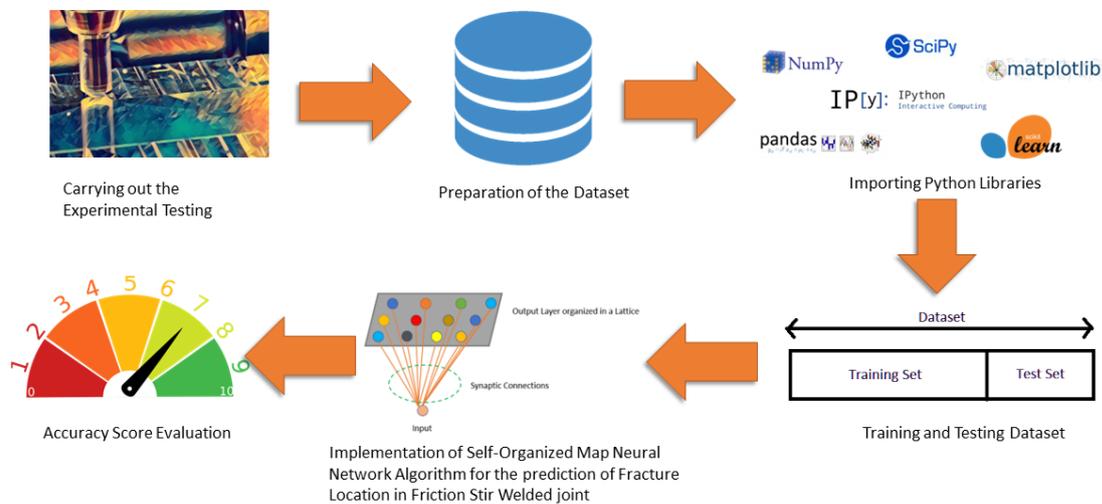

Figure 1: Flow chart of the Methodology

The third step is to import the Python libraries such as Pandas, NumPy, matlplotlib, scipy and sklearn for the computation purpose. One of the most popular tools for data cleaning and analysis in machine learning and data science is called Pandas. Pandas is the ideal tool to deal with this chaotic real-world data in this situation. And one of the accessible Python packages constructed on top of NumPy is pandas. Numerous mathematical operations can be carried out on arrays with NumPy. It provides a vast library of high-level arithmetic operations that work on these arrays and matrices, as well as strong data structures that facilitate optimal computation with arrays and matrices. A well-liked Python library for visualization techniques is Matplotlib. It is not primarily linked with machine learning, like Pandas. It is especially helpful when a coder needs to see the data's patterns. It is a library for 2D charting used to produce 2D graphs and maps. The given data is subjected to mathematical processes using the scipy special functions. A module for Scipy's special functions is included in the Scipy package. The most practical Platform for machine learning is definitely scikit-learn. Numerous effective methods for machine learning and statistical modeling, such as classification, regression, clustering, and dimensionality reduction, are included in the Sklearn package. The fourth step is to split the dataset into testing and training sets i.e., 80% of data is used for training purposes and 20% of data is used for testing purposes. The last step is to subject the Self-Organizing Maps Neural Network and further evaluate the accuracy score on the used dataset.

### 3. Results and Discussion

Self-organizing maps is an unsupervised machine learning algorithm that works on the principle of competitive learning. There is a special organization in the distribution of neurons. So here we are basically talking in terms of the lattice of the output neurons which can be arranged as a one-dimensional lattice, two-dimensional lattice, or even a higher dimensional lattice space where the neurons will be organized. Higher dimensional lattice space involves a lot of complexity, so due to this reason from a practical point of view we use one-dimensional and dimensional lattice space. If those neurons are connected to the inputs in some manner and the input patterns are fed which act as a stimulus to the neurons which are present at outputs. When the stimulus is present then out of different neurons that are existing in the lattice then one of them will be the winner and the synaptic connections from the input layer to the output layer will be adjusted and then weight updating takes place in such a way that the Euclidian distance between the input vector and the weight vector is minimized. The minimization of the Euclidian distance means the maximization of the $w^T x$ which is the output. It is noted that one of the neurons emerges as a winner and the different patterns are fed through the input space to the given system. Input distribution is basically a non-uniform distribution but if we start with a regular lattice structure depending on the input statistics then the ultimate organization of the lattice structure will be slightly different, and these results will be indicative of the statistics of the input patterns which are applied as stimuli. It should be noted that the neurons which are present at the output act in a competitive manner this means that they inhibit the responses from each other. From a neurobiological point of view, it is observed that the neurons which are closer to the winning neurons have an excitatory response. On the other hand, inhibitor response is created by the neurons which are far from the way from the winning neurons. This means that the Self Organizing Maps network will exhibit short-range excitation and long-range inhibition. A self-organizing map is a neurobiologically motivated algorithm. There are two models of self-organizing maps. i.e., von-der Malsburg model and other more general model is Kohnen Model. But in the recent study, we have used the Kohnen Model. In Kohnen Model, we generally take an output model where the output layer is organized in a lattice. Input need not be connected in the form of lattice and this input will be connected to all the outputs which will constitute a bundle of synaptic connections as shown in Figure 2. Half of the data compression is possible in this case because a number of inputs is less than those of output. Kohnen model is basically based on a vector coding algorithm which optimally places a fixed number of vectors into a higher dimensioning input space.

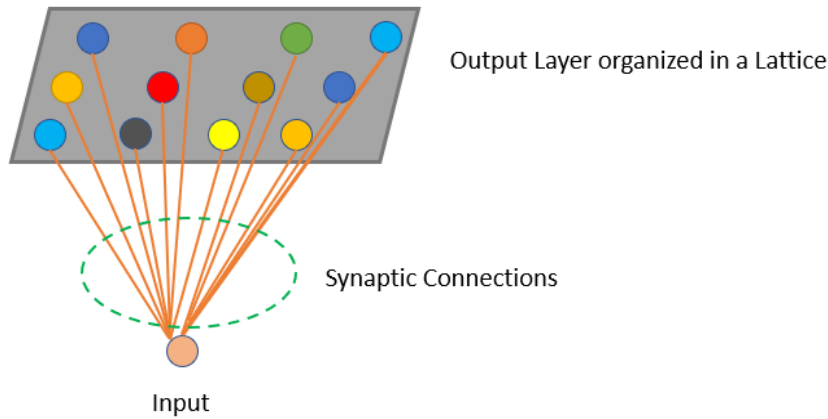

Figure 2: Kohnen Model of Self-Organizing Maps Neural Network in the present study

There are some essential processes that need to be fulfilled for self-organizing maps. The first process is called Competition which means that for each input pattern the neurons present in the output layer will determine a discriminant function that provides a basis for the competition. The particular neuron with the largest discriminant function emerges as the winner Second process is cooperation, in which the winning neurons determine the topological neighborhood of the excited neuron. The excitation process is the cooperation that not only strengthens the winning neuron but also strengthens the neurons which are closer to the winning neurons. So, it should be noted that competition refers to the long-range inhibition process while cooperation refers to the short-range excitation process. The third process is called the step of synaptic adaptation which enables the excited neurons to increase their discriminant function in response to the stimulus which has caused the winning of the neuron.

In competitive process we considered m dimensional input so that the input $\vec{x}$ is given by the equation 1.

$$\vec{x} = [x_1 \quad x_2 \quad ............ x_n]^T \qquad (1)$$

The weight is given by the equation 2.

$$\vec{w_j} = [w_{j1} \quad w_{j2} ......w_{jm}]^T \quad , j=1,2,3....l \qquad (2)$$

Where l is the number of output neurons. It should be noted that every output is connected to the inputs. The main objective is to determine the best match between $\vec{x}$ and $\vec{w_j}$. So there will be competition between l number of output neurons in which $\vec{x}$ is going to find out the best match with $\vec{w_j}$ which emerges as the winner and in this case the winning index will be j and corresponding weight will be the winning weight vector. The maximization process is represented in equation 3 which leads to the minimization of Euclidean Distance as shown in equation 4.

$$Max \ |\vec{w_j}^T \cdot \vec{x}| \qquad (3)$$

$$Min \ \|\vec{x} - \vec{w_j}\| \qquad (4)$$

Index of the winning neuron can be processed by using equation 5.

$$i(\vec{x}) = \arg \min_j \|\vec{x} - \vec{w_j}\| \qquad (5)$$

We use majority voting to allocate a single label per each neuron on the map in order to create a label map. If a neuron is chosen without a Best Matching Unit (BMU), The final iteration's label map is shown in Figure 3. A large number of neurons neither are 0 nor 1 at the beginning, and the class labels seem to be dispersed haphazardly. The last iteration clearly separates the class 0 and 1 region; however, we still notice a few cells that are neither in the final iteration.

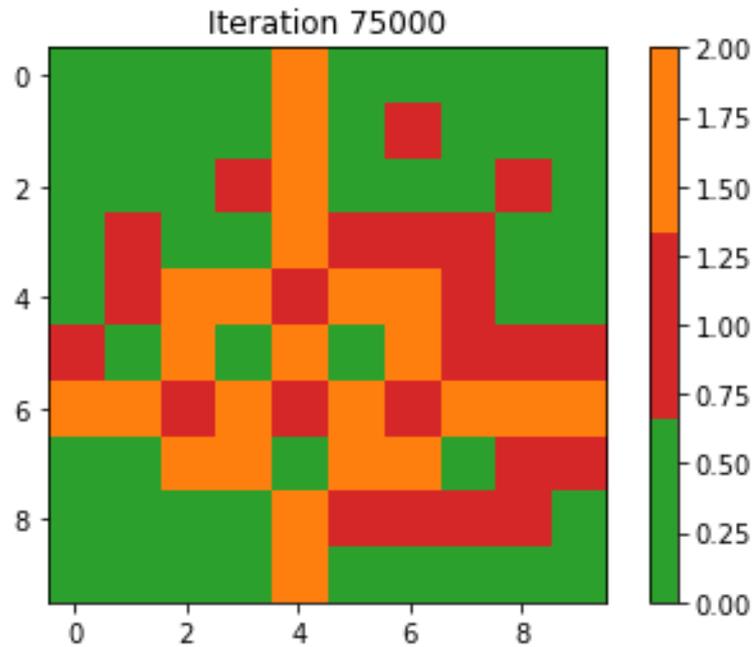

Figure 3: Labled Map obtained at final iteration

The accuracy score obtained as result was 0.9692. The dataset we used for this study is clean, with a relatively small number of features and observations. The data scientists encounter far more complex problems in the real world, and the labeled dataset is not entirely available. Although if they are offered, their quality might not be trustworthy.

## 4. Conclusions

The implementation of an unsupervised Self-Organizing Map Neural Network algorithm for a classification problem is shown in the current paper. We used a data without labels to train the map, and we projected the labels onto the map to verify the training's outcome. As anticipated, we could see that each class has distinct regions, with neurons with related characteristics clustered closer together. Finally, we evaluated the map's prediction accuracy using an unanticipated test dataset. Implementing the Self-Organizing Map Neural Network to a real-world situation presents certain difficulties for us. First, without a tagged dataset, we are unable to calculate the loss. We have no way to verify the dependability of the trained map. The features of the data themselves have a significant impact on the map's quality. The distance-based approach requires the normalization of the data as a preprocessing step. Understanding the distribution of the data points necessitates an initial analysis of the dataset. We may employ other dimensionality reduction techniques, such as PCA and Singular Value Decomposition, especially for data with high dimensionality where visualization is not possible (SVD).